\documentclass{article}

\PassOptionsToPackage{numbers, compress}{natbib}

\usepackage[preprint]{neurips_2020}

\usepackage[utf8]{inputenc} 
\usepackage[T1]{fontenc}    
\usepackage[pagebackref=true,breaklinks=true,letterpaper=true,colorlinks,bookmarks=false]{hyperref}
\usepackage{url}            
\usepackage{booktabs}       
\usepackage{amsfonts}       
\usepackage{nicefrac}       
\usepackage{microtype}      

\usepackage{amssymb}
\usepackage{latexsym}
\usepackage{amsmath}
\usepackage{amsthm}
\usepackage{graphicx}
\usepackage{subfigure}
\usepackage{caption}

\newcommand\blfootnote[1]{%
  \begingroup
  \renewcommand\thefootnote{}\footnote{#1}%
  \addtocounter{footnote}{-1}%
  \endgroup
}

\DeclareMathOperator*{\argmin}{argmin}

\title{Differentiable Forward and Backward \\ Fixed-Point Iteration Layers}

\author{%
  Younghan Jeon$^{\dagger*}$ \hspace{5em} Minsik Lee$^{\ddagger*}$ \hspace{5em} Jin Young Choi$^{\dagger}$ \\
  \texttt{yh1992@snu.ac.kr} \hspace{1em} \texttt{mleepaper@hanyang.ac.kr} \hspace{1em} \texttt{jychoi@snu.ac.kr}\\
  Department of Electrical and Computer Engineering, ASRI, Seoul National University$^{\dagger}$ \\
  Division of Electrical Engineering, Hanyang University$^{\ddagger}$ \\
}

\begin{document}

\maketitle

\begin{abstract}

Recently, several studies proposed methods to utilize some classes of optimization problems in designing deep neural networks to encode constraints that conventional layers cannot capture. However, these methods are still in their infancy and require special treatments, such as analyzing the KKT condition, for deriving the backpropagation formula. In this paper, we propose a new layer formulation called the fixed-point iteration (FPI) layer that facilitates the use of more complicated operations in deep networks. The backward FPI layer is also proposed for backpropagation, which is motivated by the recurrent back-propagation (RBP) algorithm. But in contrast to RBP, the backward FPI layer yields the gradient by a small network module without an explicit calculation of the Jacobian. In actual applications, both the forward and backward FPI layers can be treated as nodes in the computational graphs. All components in the proposed method are implemented at a high level of abstraction, which allows efficient higher-order differentiations on the nodes.
In addition, we present two practical methods of the FPI layer, FPI\_NN and FPI\_GD, where the update operations of FPI are a small neural network module and a single gradient descent step based on a learnable cost function, respectively. FPI\_NN is intuitive, simple, and fast to train, while FPI\_GD can be used for efficient training of energy networks that have been recently studied. 
While RBP and its related studies have not been applied to practical examples, our experiments show the FPI layer can be successfully applied to real-world problems such as image denoising, optical flow, and multi-label classification.
\end{abstract}

\blfootnote{$^*$ Authors contributed equally.}

\section{Introduction}
\label{intro}




Recently, several papers proposed to compose a deep neural network with more complicated algorithms rather than with simple operations as it had been used.
For example, there have been methods using certain types of optimization problems in deep networks such as differentiable optimization layers \cite{optnet} and energy function networks \cite{icnn, spen}. 
These methods can be used to introduce a prior in a deep network and provide a possibility of bridging the gap between deep learning and some of the traditional methods. However, they are still premature and require non-trivial efforts to implement in actual applications. Especially, the backpropagation formula has to be derived explicitly for each different formulation based on some criteria like the KKT conditions, etc. This limits the practicality of the approaches since there can be numerous different formulations depending on the actual problems.

Meanwhile, there has been an algorithm called recurrent back-propagation (RBP) proposed by Almeida \cite{almeida} and Pineda \cite{pineda} several decades ago. RBP is a method to train an RNN that converges to the steady state. The advantages of RBP are that it can be applied universally to most operations that consist of repeated computations and that the whole process can be summarized by a single update equation. Even with its long history, however, RBP and related studies \cite{deq, reviving} have been tested only for verifying the theoretical concept and there has been no example that applied these methods to a practical task. Moreover, there have been no studies using RBP in conjunction with other neural network components to verify the effect in more complex settings.

In this paper, to facilitate the use of more complicated operations in deep networks, we introduce a new layer formulation that can be practically implemented and trained based on RBP with some additional considerations. To this end, we employ the fixed-point iteration (FPI), which is the basis of many numerical algorithms including most gradient-based optimizations, as a \textit{layer} of a neural network.
In the FPI layer, the layer's input and its weights are used to define an update equation, and the output of the layer is the fixed-point of the update equation. Under mild conditions, the FPI layer is differentiable and the derivative depends only on the fixed point, which is much more efficient than adding all the individual iterations to the computational graph.

We also propose a backpropagation method called \emph{backward FPI layer} based on RBP \cite{almeida, pineda} to compute the derivative of the FPI layer efficiently. We prove that if the aforementioned conditions for the FPI layer hold, then the backward FPI layer also converges. In contrast to RBP, the backward FPI layer yields the gradient by a small network module which allows us to avoid the explicit calculation of the Jacobian. 
In other words, we do not need a separate derivation for the backpropagation formula and can utilize existing autograd functionalities. Especially, we provide a modularized implementation of the partial differentiation operation, which is essential in the backward FPI layer but is absent in regular autograd libraries, based on an independent computational graph. This makes the proposed method very simple to apply to various practical applications.
Since FPI covers many different types of numerical algorithms as well as optimization problems, there are a lot of potential applications for the proposed method.
FPI layer is highly modularized so it can be easily used together with other existing layers such as convolution, ReLU, etc.,\footnote{The code will be available upon publication.} and has a richer representation power than the feedforward layer with the same number of weights.
Contributions of the paper are summarized as follows.

\begin{itemize}
\item We propose a method to use an FPI as a layer of neural network. The FPI layer can be utilized to incorporate the mechanisms of conventional iterative algorithms, such as numerical optimization, to deep networks. Unlike other existing layers based on differentiable optimization problems, the implementation is much simpler and the backpropagation formula can be universally derived.
\item For backpropagation, the backward FPI layer is proposed based on RBP to compute the gradient efficiently.
Under the mild conditions, we show that both forward and backward FPI layers are guaranteed to converge.
All components are highly modularized and a general partial differentiation tool is developed so that the FPI layer can be used in various circumstances without any modification.
\item Two types of FPI layers (FPI\_NN, FPI\_GD) are presented. 
The proposed networks based on the FPI layers are applied to practical tasks such as image denoising, optical flow, and multi-label classification, 
which have been largely absent in existing RBP-based studies, and show good performance.
\end{itemize}

The remainder of this paper is organized as follows: We first introduce related works in Section \ref{work}. The proposed FPI layer is explained in Section \ref{method}, the experimental results follow in Section \ref{expr}. Finally, we conclude the paper in Section \ref{conc}.
\section{Background and Related Work}
\label{work}

\textbf{Fixed-point iteration:} For a given function $g$ and a sequence of vectors, $\{ x_n \in \mathbb{R}^d \}$, the fixed-point iteration \cite{fpibook} is defined by the following update equation
\begin{equation}
\label{eq_fpi_base}
    x_{n+1} = g(x_{n}),~~ n = 0, 1, 2, \cdots,
\end{equation}
that converges to a fixed point $\hat{x}$ of $g$, satisfying $\hat{x}=g(\hat{x})$. 
The gradient descent method ($x_{n+1}=x_{n}-\gamma\nabla f(x_{n})$) 
is a popular example of fixed-point iteration. Many numerical algorithms are based on fixed-point iteration, and there are also many examples in machine learning.
Here are some important concepts about fixed-point iteration.

\textit{Definition 1 (Contraction mapping)} \cite{contraction}. On a metric space $(X, d)$, the function $f:X\rightarrow X$ is a contraction mapping if there is a real number $0\leq k<1$ that satisfies the following inequality for all $x_1$ and $x_2$ in $X$.
\begin{equation}
\label{eq_contraction}
    d\big(f(x_1), f(x_2)\big) \leq k \cdot d(x_1, x_2).
\end{equation}
The smallest $k$ that satisfies the above condition is called the Lipschitz constant of $f$.
The distance metric is defined to be an arbitrary norm $\Vert \cdot \Vert$ in this paper.
Based on the above definition, the Banach fixed-point theorem \cite{banach} states the following.

\textit{Theorem 1 (Banach fixed-point theorem).} \textit{A contraction mapping has exactly one fixed point and it can be found by starting with any initial point and iterating the update equation until convergence.}

Therefore, if $g$ is a contraction mapping, it converges to a unique point $\hat{x}$ regardless of the starting point $x_0$. The above concepts are important in deriving the proposed FPI layer in this paper.

\textbf{Energy function networks:}
Scalar-valued networks to estimate the energy (or error) functions have recently attracted considerable research interests.
The energy function network (EFN) has a different structure from general feed-forward neural networks, and the concept was first proposed in \cite{ebl}.
After training an EFN for a certain task, the answer to a test sample is obtained by finding the input of the trained EFN that minimizes the network's output.
The structured prediction energy network (SPEN) \cite{spen} performs a gradient descent on an energy function network to find the solution, and a structured support vector machine \cite{ssvm} loss is applied to the obtained solution.
The input convex neural networks (ICNNs) \cite{icnn} are defined in a specific way so that the networks have convex structures with respect to (w.r.t.) the inputs, and their learning and inference are performed by the entropy method which is derived based on the KKT optimality conditions.
The deep value networks \cite{dvn} and the IoU-Net \cite{iounet} directly learn the loss metrics such as the intersection over union (IoU) of bounding boxes and then perform inference by gradient-based optimization methods.

Although the above approaches provide novel ways of utilizing neural networks in optimization frameworks, they have not been combined with other existing deep network components to verify their effects in more complicated problems.
Moreover, they are mostly limited to a certain type of problems and require complicated learning processes.
Our method can be applied to broader situations than EFN approaches, and these approaches can be equivalently implemented by the proposed method once the update equation for the optimization problem is derived.

\textbf{Differentiable optimization layers:}
Recently, a few papers using optimization problems as a layer of a deep learning architecture have been proposed.
Such a structure can contain a more complicated behavior in one layer than the usual layers in neural networks, and can potentially reduce the depth of the network.
OptNet \cite{optnet} presents how to use the quadratic program (QP) as a layer of a neural network.
They also use the KKT conditions to compute the derivative of the solution of QP.
Agrawal et al. \cite{dcol} propose an approach to differentiate disciplined convex programs which is a subclass of convex optimization problems.
There are a few other researches trying to differentiate optimization problems such as submodular models \cite{submodular}, cone program \cite{conic}, semidefinite program \cite{satnet}, and so on.
However, most of them have limited applications and users need to adapt their problems to the rigid problem settings.
On the other hand, our method makes it easy to use a large class of iterative algorithms as a network layer, which also includes the differentiable optimization problems.

\textbf{Recurrent back-propagation:}
RBP is a method to train a special case of RNN proposed by Almeida \cite{almeida} and Pineda \cite{pineda}.
RBP computes the gradient of the steady state for an RNN with constant memory.
Although RBP has great potential, it is rarely used in practical problems of deep learning.
Some artificial experiments showing its memory efficiency were performed, but it was difficult to apply in complex and practical tasks.
Recently, Liao et al. \cite{reviving} tried to revive RBP using the conjugate gradient method and the Neumann series.
However, both the forward and backward passes use a fixed number of steps (maximum 100), which might not be sufficient for convergence in practical problems.
Also, if the forward pass does not converge, the equilibrium point is meaningless so it can be unstable to train the network using the unconverged final point, which is a problem not addressed in the paper.
Deep equilibrium models (DEQ) \cite{deq} tried to find the equilibrium points of a deep sequence model via an existing root-finding algorithm.
Then, for back-propagation, they compute the gradient of the equilibrium point by another root-finding method. In short, both the forward and backward passes are implemented via quasi-Newton methods. DEQ can also be performed with constant memory, but it can only model the sequential (temporal) data, and the aforementioned convergence issues still exist.

RBP-based methods mainly perform experiments to verify the theoretical concepts and have not been well applied to practical examples. Our work incorporates the concept of RBP in the FPI layer to apply complicated iterative operations in deep networks, and presents two types of algorithms accordingly. The proposed method is the first RBP-based method that shows successful applications to practical tasks in machine learning or computer vision, and can be widely used for promotion of the RBP-based research in the deep learning field.


\section{Proposed Method}
\label{method}

The fixed-point iteration formula contains a wide variety of forms and can be applied to most iterative algorithms.
Section \ref{structure} describes the basic structure and principles of the FPI layer.
Section \ref{diff} and \ref{backward} explains the differentiation of the layer for backpropagation.
Section \ref{convergence} describes the convergence of the FPI layer.
Section \ref{application} presents two exemplar cases of the FPI layer.

\subsection{Structure of the fixed-point iteration layer}
\label{structure}
Here we describe the basic operation of the FPI layer.
Let $g(x, z; \theta)$ be a parametric function where $x$ and $z$ are vectors of real numbers and $\theta$ is the parameter. We assume that $g$ is differentiable for $x$ and also has a Lipschitz constant less than one for $x$, and the following fixed point iteration converges to a unique point according to the Banach fixed-point iteration theorem:
\begin{equation}
\label{eq_fpi}
    x_{n+1} = g(x_{n}, z; \theta),
  \quad   \hat{x} = \lim_{n\to\infty} x_{n}
\end{equation}
The FPI layer can be defined based on the above relations. The FPI layer $\mathbf{F}$ receives an observed sample or output of the previous layer as input $z$, and yields the fixed point $\hat{x}$ of $g$ as the layer's output, i.e.,
\begin{equation}
\label{eq_fpi_layer}
    \hat{x} = g(\hat{x}, z; \theta) = \mathbf{F}(x_{0}, z;\theta)=
    \lim_{n\to\infty} g^{(n)}(x_0, z; \theta) = (g \circ g \circ \cdots \circ g)(x_0, z; \theta)
\end{equation}
where $\circ$ indicates the function composition operator. 
The layer receives the initial point $x_0$ as well, but its actual value does not matter in the training procedure because $g$ has a unique fixed point. 
Hence, $x_0$ can be predetermined to any value such as zero matrix. Accordingly, we will often express $\hat{x}$ as a function of $z$ and $\theta$, i.e., $\hat{x}(z;\theta)$.
When using an FPI layer, the first equation in (\ref{eq_fpi}) is repeated until convergence to find the output $\hat{x}$. We may use some acceleration techniques such as the Anderson acceleration \cite{anderson} for faster convergence.
Note that there is no apparent relation between the shapes of $x_n$ and $z$. Hence the sizes of the input and output of an FPI layer do not have to be same.

\subsection{Differentiation of the FPI layer}
\label{diff}

Similar to other network layers, learning of $\mathbf{F}$ is performed by updating $\theta$ based on backpropagation. For this, the derivatives of the FPI layer has to be calculated. 
One simple way to compute the gradients is to construct a computational graph for all the iterations up to the fixed point $\hat{x}$.
However, this method is not only time consuming but also requires a lot of memory.

In this section, we show that the derivative of the entire FPI layer depends only on the fixed point $\hat{x}$. In other words, all the $x_n$ before convergence are actually not needed in the computation of the derivatives. Hence, we can only retain the value of $\hat{x}$ to perform backpropagation, and consider the entire $\mathbf{F}$ as a node in the computational graph.
Note that $\hat{x} = g(\hat{x}, z; \theta)$ is satisfied at the fixed point $\hat{x}$.
If we differentiate both sides of the above equation w.r.t. $\theta$, we have
\begin{equation}
\label{eq_dxhat_1}
\begin{split}
    \frac{\partial \hat{x}}{\partial \theta} 
    & = \frac{\partial g}{\partial  \theta}(\hat{x}, z; \theta) + \frac{\partial g}{\partial x}(\hat{x}, z; \theta)    ~\frac{\partial \hat{x}}{\partial  \theta}.
\end{split}
\end{equation}
Here, $z$ is not differentiated because $z$ and $\theta$ are independent. Rearranging the above equation gives
\begin{equation}
\label{eq_dxhat_3}
    \cfrac{\partial \hat{x}}{\partial \theta} = \left( I - \cfrac{\partial g}{\partial x}(\hat{x}, z; \theta)\right) ^{-1}\cfrac{\partial g}{\partial \theta}(\hat{x}, z; \theta),
\end{equation}
which confirms the fact that the derivative of the output of $\mathbf{F}(x_0, z; \theta) = \hat{x}$ depends only on the value of $\hat{x}$.
One downside of the above derivation is that it requires the calculation of Jacobians of $g$, which may need a lot of memory space (e.g., convolutional layers). Moreover, calculating the inverse can also be a burden. In the next section, we will provide an efficient way to resolve these issues.

\subsection{Backward fixed-point iteration layer}
\label{backward}
To train the FPI layer, we need to obtain the gradient w.r.t. its parameter $\theta$. In contrast to RBP \cite{almeida, pineda}, we propose a computationally efficient layer, called the backward FPI layer, that yields the gradient without explicitly calculating the Jacobian.
Here, we assume that an FPI layer is in the middle of the network. If we define the loss of the entire network as $L$, then what we need for backpropagation of the FPI layer is $\nabla_{\theta} L(\hat{x})$.
According to (\ref{eq_dxhat_3}), we have
\begin{equation}
\label{eq_gL}
    \nabla_{\theta} L = \left(\cfrac{\partial \hat{x}}{\partial \theta}\right)^\top\nabla_{\hat{x}}L 
    = \left(\cfrac{\partial g}{\partial\theta}(\hat{x}, z; \theta)\right)^\top\left( I - \cfrac{\partial g}{\partial x}(\hat{x}, z; \theta)\right) ^{-\top}\nabla_{\hat{x}}L.
\end{equation}
This section describes how to calculate the above equation efficiently.
(\ref{eq_gL}) can be divided into two steps as follows:
\begin{equation}
\label{eq_back1}
    c = \left( I - \cfrac{\partial g}{\partial x}(\hat{x}, z; \theta)\right)^{-\top}\nabla_{\hat{x}}L, \quad
      \nabla_{\theta}L = \left(\cfrac{\partial g}{\partial\theta}(\hat{x}, z; \theta)\right)^\top c.
\end{equation}
Rearranging the first equation in (\ref{eq_back1}) yields $c = \left(\cfrac{\partial g}{\partial x}(\hat{x}, z; \theta)\right)^\top c + \nabla_{\hat{x}}L$, which 
can be expressed as an iteration form, i.e., 
\begin{equation}
\label{eq_back}
    c_{n+1} = \left(\cfrac{\partial g}{\partial x}(\hat{x}, z; \theta)\right)^\top c_{n} + \nabla_{\hat{x}}L,
\end{equation}
which corresponds to RBP. This iteration eliminates the need of the inverse calculation but it still requires the calculation of the Jacobian of $g$ w.r.t. $\hat{x}$.
Here, we derive a new network layer, i.e. the backward FPI layer, that yields  
the gradient without an explicit calculation of the Jacobian.
To this end, we define a new function $h$ as $ h(x, z, c; \theta) = c^{\top}g(x, z; \theta),$
then (\ref{eq_back}) becomes
\begin{equation}
\label{eq_back_h}
    c_{n+1} = \cfrac{\partial h}{\partial x}(\hat{x}, z, c_{n}; \theta) + \nabla_{\hat{x}}L.
\end{equation}
Note that the output of $h$ is scalar. Here, we can consider $h$ as another small network containing only a single step of $g$ (with an additional inner product).
The gradient of $h$ can be computed based on existing autograd functionalities with some additional considerations.
Similarly, the second equation in (\ref{eq_back1}) is expressed using $h$:
\begin{equation}
\label{eq_back2_h}
    \nabla_{\theta}L = \cfrac{\partial h}{\partial\theta}(\hat{x}, z, c; \theta),
\end{equation}
where $c$ is the fixed point obtained from the fixed-point iteration in (\ref{eq_back_h}).
In this way, we can compute $\nabla_{\theta}L$ by (\ref{eq_back2_h}) without any memory-intensive operations and Jacobian calculation. 
$c$ can be obtained by initializing $c$ to some arbitrary value and repeating the above update until convergence.

Note that this backward FPI layer can be treated as a node in the computational graph, hence the name backward FPI \emph{layer}.
However, care should be taken about the above derivation in that the differentiations w.r.t. $x$ and $\theta$ are \emph{partial} differentiations. $x$ and $\theta$ might have some dependency with each other, which can disrupt the partial differentiation process if it is computed based on a usual autograd framework. Let $\phi(a, b)$ hereafter denotes the gradient operation in the conventional autograd framework that calculates the derivative of $a$ w.r.t $b$ where $a$ and $b$ are both nodes in a computational graph. Here, $b$ can also be a set of nodes, in which case the output of $\phi$ will also be a set of derivatives.

In order to resolve the issue, we implemented a general partial differentiation operator $P(s; r) \triangleq \partial r(s) / \partial s$ where $s$ is a set of nodes, $r$ is a function (a function object, to be precise), and $\partial r(s) / \partial s$ denotes the set of corresponding partial derivatives: Let $I(t)$ denotes an operator that creates a set of leaf nodes by cloning the nodes in the set $t$ and detaching them from the computational graph.
$P$ first creates an independent computational graph having leaf nodes $s'=I(s)$. These leaf nodes are then passed onto $r$ to yield $r' = r(s')$, and now we can differentiate $r'$ w.r.t. $s'$ using $\phi(r', s')$ to calculate the partial derivatives, because the nodes in $s'$ are independent to each other. Here, the resulting derivatives $\partial r' / \partial s'$ are also attached in the independent graph as the output of $\phi$. The $P$ operator creates another set of leaf nodes $I(\partial r' / \partial s')$, which is then attached to the original graph (where $s$ resides) as the output of $P$, i.e., $\partial r(s) / \partial s$. In this way, the whole process is completed and the partial differentiation can be performed accurately. If some of the partial derivatives are not needed in the process, we can simply omit them in the calculation of $\partial r' / \partial s'$.

Note that the above independent graph is preserved for backpropagation. Let $H(v; u)$ be an operator that creates a new function object that calculates $\sum_i \left<v_i, u_i\right>$, where the node $v_i$ is an element of the set $v$ and $u_i$ is one of the outputs of the function object $u$. In the backward path of $P$, the set $\delta$ of gradients passed onto $P$ by backpropagation is used to create a function object $\eta=H(\delta; \rho)$ where $\rho(s)$ is a function object that calculates $P(s; r) = \partial r(s) / \partial s$. Then, the backpropagated gradients for $s$ can be calculated with another $P$ operation, i.e., $P(\delta \cup s; \eta)$ (here, the derivatives for $\delta$ do not need to be calculated). In practice, the independent graph created in the forward path is reused for $\rho$ in the calculation of $\eta$.

The backward FPI layer can be highly modularized with the above operators, i.e., $P$, $H$, and a plus operator can construct (\ref{eq_back_h}) and (\ref{eq_back2_h}) entirely, and the iteration of (\ref{eq_back_h}) can be implemented with another forward FPI layer. This allows multiple differentiations of the forward and backward FPI layers. A picture depicting all the above processes is provided in the supplementary material.
All the forward and backward layers are implemented at a high level of abstraction, and therefore, it can be easily applied to practical tasks by changing the structure of $g$ to one that is suitable for each task.

\subsection{Convergence of the FPI layer}
\label{convergence}

The forward path of the FPI layer converges if the bounded Lipschitz assumption holds.
For example, to make a fully connected layer a contraction mapping, simply dividing the weight by a number greater than the maximum singular value of the weight matrix 
will suffice.
In practice, we empirically found out that setting the initial values of weights ($\theta$) to small values is enough for making $g$ a contraction mapping throughout the training procedure.

\textbf{Convergence of the backward FPI layer.}
The backward FPI layer is composed of a linear mapping based on the Jacobian $\partial{g}/\partial{x}$ on $\hat{x}$.
Convergence of the backward FPI layer can be confirmed by the following proposition.

\textbf{Proposition 1.}
\textit{If $g$ is a contraction mapping, the backward FPI layer (Eq. (\ref{eq_back})) converges to a unique point.}
\begin{proof}
\vspace{-0.4cm}
For simplicity, we omit $z$ and $\theta$ from $g$. By the definition of the contraction mapping and the assumption of the arbitrary norm metric, $\frac{\| g(x_2) - g(x_1) \|}{\| x_2 - x_1 \|} \leq k$ is satisfied
for all $x_1$ and $x_2$ ($0 \leq k < 1$).
For a unit vector $v$, i.e., $\|v\| = 1$ for the aforementioned norm, and a scalar $t$, let $x_2 = x_1 + tv$.
Then, the above inequality becomes $\frac{\| g(x_1 + tv) - g(x_1) \|}{\| t \|} \leq k$.
For another vector $u$ with $\|u\|_* \le 1$ where $\| \cdot \|_*$ indicates the dual norm, it satisfies
\begin{equation}
\label{eq_pf3}
    \cfrac{u^\top\big(g(x_1 + tv) - g(x_1)\big)}{\vert t \vert} \leq \cfrac{\Vert g(x_1 + tv) - g(x_1)) \Vert}{\vert t \vert} \leq k 
\end{equation}
based on the definition of the dual norm. This indicates that
\begin{equation}
\label{eq_pf4}
    \lim_{t\to0^+}\cfrac{u^\top\big(g(x_1 + tv) - g(x_1)\big)}{\vert t \vert} = \nabla_{v}(u^{\top}g)(x_1) \leq k.
\end{equation}
According to the chain rule, $\nabla(u^{\top}g) = \big(u^{\top}J_g\big)^{\top}$ where $J_g$ is the Jacobian of $g$. That gives
\begin{equation}
\label{eq_pf5}
    \nabla_{v}(u^{\top}g)(x_1)
    = \big(\nabla(u^{\top}g)(x_1)\big)^{\top}\cdot v
    = u^{\top}J_g(x_1)~v \leq k.
\end{equation}
Let $x_1 = \hat{x}$ then $u^{\top}J_g(\hat{x})~v \leq k$ for all $u$, $v$ that satisfy $\Vert u\Vert_*=\Vert v\Vert=1$.
Therefore,
\begin{equation}
\label{eq_pf6}
    \Vert J_g(\hat{x})\Vert = \sup_{\|v\|=1} \|J_g(\hat{x}) v\| = \sup_{\|v\|=1, \|u\|_* \le 1} u^\top J_g(\hat{x}) v \leq k < 1
\end{equation}
which indicates that the linear mapping by weight $J_g(\hat{x})$ is a contraction mapping.
By the Banach fixed-point theorem, the backward FPI layer converges to the unique fixed-point.
\end{proof}

\subsection{Two representative cases of the FPI layer}
\label{application}
As mentioned before, FPI can take a wide variety of forms.
We present two representative methods that are easy to apply to practical problems.
\subsubsection{Neural net FPI layer (FPI\_NN)}
The most intuitive way to use the FPI layer is to set $g$ to an arbitrary neural network module. In FPI\_NN, the input variable recursively enters the same network module until convergence.
$g$ can be composed of layers that are widely used in deep networks such as convolution, ReLU, and linear layers.
FPI\_NN can perform more complicated behaviors with the same number of parameters than using $g$ directly without FPI, as demonstrated in the experiments section.

\subsubsection{Gradient descent FPI layer (FPI\_GD)}
The gradient descent method can be a perfect example for the FPI layer. This can be used for efficient implementations of the energy function networks like ICNN \cite{icnn}.
Unlike a typical network which obtains the answer directly as the output of the network (i.e., $f(a; \theta)$ is the answer of the query $a$), an energy function network retrieves the answer by optimizing an input variable of the network (i.e., $\argmin_{x}f(x, a; \theta)$ becomes the answer).
The easiest way to optimize the network $f$ is gradient descent ($x_{n+1}=x_{n}-\gamma\nabla f(x_{n}, a; \theta)$).
This is a form of FPI and the fixed point $\hat{x}$ is the optimal point of $f$, i.e., $\hat{x} = \argmin_{x}f(x, a; \theta)$.
In case of a single FPI layer network with FPI\_GD, $\hat{x}$ becomes the final output of the network. Accordingly, this output is fed into the final loss function $L\big(x^*, \hat{x})$
to train the parameter $\theta$ during the training procedure.
This behavior conforms to that of an energy function network.
However, unlike the existing methods, the proposed method can be trained easily with the universal backpropagation formula. Therefore, the proposed FPI layer can be an effective alternative to train energy function networks.
An advantage of FPI\_GD is that it can easily satisfy the bounded Lipschitz condition by adjusting the step size $\gamma$.

\section{Experiments}
\label{expr}
Since several studies \cite{almeida, pineda, reviving, deq} have already shown that RBP-based algorithms require only a constant amount of memory, we omit memory-related experiments.
Instead, we focus on applying the proposed method to practical tasks in this paper.
It is worth noting that both the forward and backward FPI layers were highly modularized, and the exactly same implementations were shared across all the experiments without any alteration. The only difference was the choice of $g$ where we could simply plug in its functional definition, which shows the efficiency of the proposed framework.
In the image denoising experiment, we compare the performance of the FPI layer to a non-FPI network that has the same structure as $g$.
In the optical flow problem, a relatively very small FPI layer is attached at the end of FlowNet \cite{flownet} to show its effectiveness.
For all experiments, the detailed structure of $g$ and the hyperparameters for training are described in the supplementary material.
Results on the multi-label classification problem show that the FPI layer is superior in performance compared to the existing state-of-the-art algorithms.
All training was performed using the Adam \cite{adam} optimizer.

\subsection{Image denoising}

Here, we compare the image denoising performance for gray images perturbed by Gaussian noise with variance $\sigma^2$. Image denoising has been traditionally solved with iterative, numerical algorithms, hence using an iterative structure like the proposed FPI layer can be an appropriate choice for the problem.
To generate the image samples, we cropped the images in the Flying Chairs dataset \cite{flownet} and converted it to gray scale (400 images for training and 100 images for testing).
We constructed a single FPI\_NN layer network for this experiment.
For comparison, we also constructed a feedforward network that has same structure as $g$.
The performance is reported in terms of peak signal-to-noise ratio (PSNR) in Table \ref{table_dn}.
\begin{table}[h]
\centering
\begin{tabular}{|l|l|l|l|}
\hline
Method          & $\sigma=15$    & $\sigma=20$    & $\sigma=25$    \\ \hline
Feedforward     & 32.18 & 30.44 & 29.09 \\ \hline
FPI\_NN          & \bf{32.43} & \bf{31.00} & \bf{29.74} \\ \hline
\end{tabular}
\vspace{0.1cm}
\caption{Denoising performance (PSNR, higher is better).}
\label{table_dn}
\vspace{-0.5cm}
\end{table}

Table \ref{table_dn} shows that the single FPI layer network outperforms the feedforward network in all experiments. Note that the performance gap between the two algorithms is larger in more noisy circumstances.
Since both the networks are trained to yield the best performance in their given settings, this confirms that a structure with repeated operations can be more suitable for this type of problem. An advantage of the proposed FPI layer here is that there is no explicit calculation of the Jacobian, which can be quite large in this image-based problem, even though there was no specialized component except the bare definition of $g$ thanks to the highly modularized nature of the layer. Examples of image denoising results are shown in the supplementary material.

\subsection{Optical flow}
Optical flow is one of the major research areas of computer vision that aims to acquire motions by matching pixels in two images.
We demonstrate the effectiveness of the FPI layer by a simple experiment, where the layer is attached at the end of FlowNet \cite{flownet}.
The Flying Chairs dataset \cite{flownet} was used with the original split which has 22,232 training and 640 test samples.
In this case, the FPI layer plays the role of post-processing. 
We attached a very simple FPI layer consisting of conv/deconv layers and recorded the average end point error (aEPE) per epoch as shown in Figure \ref{fig_epe}.
Although the number of additional parameters is extremely small (less than 0.01\%) and the computation time is nearly the same with the original FlowNet, it shows noticeable performance improvement.

\subsection{Multi-label classification}

The multi-label text classification dataset (Bibtex) was introduced in \cite{multilabel}.
The goal of the task is to find the correlation between the data and the multi-label features.
Both the data and features are binary with 1836 indicators and 159 labels, respectively.
The numbers of indicators and labels differ for each data, and that of labels is unknown during the evaluation process.
We used the same train and test split as in \cite{multilabel} and evaluated the $F_1$ scores.
Here, we used two single FPI layer networks with FPI\_GD and FPI\_NN. 
We set $g$ of FPI\_NN and $f$ of FPI\_GD to similar structures which contain one or two fully-connected layers and activation functions.
As mentioned, the detailed structures of the networks are described in supplementary materials.
\begin{table}[t]
\centering
    \begin{minipage}{.4\textwidth}
    \centering
        \includegraphics[width=\textwidth]{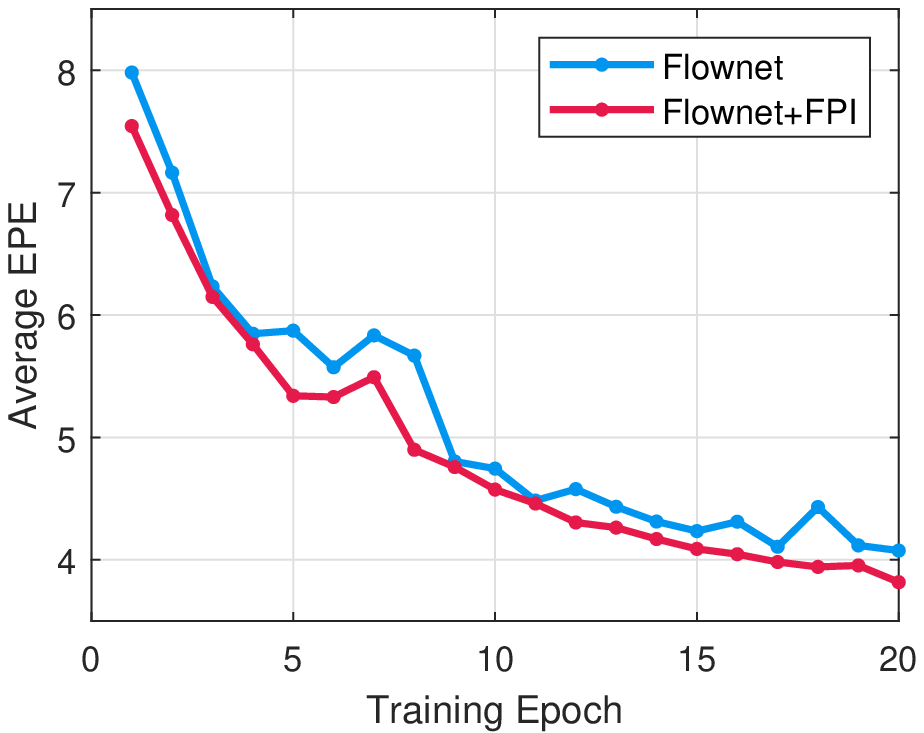}
        \vspace{0.05cm}
        \captionof{figure}{Average EPE per epoch (lower is better).}
        \label{fig_epe}
    \end{minipage}
    \hspace{0.5cm}
    \begin{minipage}{.45\textwidth}
    \centering
    \begin{tabular}{ll}
        \hline
        Method                      & F1 score \\ \hline
        MLP \cite{spen}             & 38.9     \\
        Feedforward net \cite{icnn} & 39.6     \\
        SPEN \cite{spen}            & 42.2     \\
        ICNN \cite{icnn}            & 41.5     \\
        DVN(GT) \cite{dvnurl}       & 42.9     \\
        DVN(Adversarial)\cite{dvn}  & \bf{44.7}\\ \hline
        FPI\_GD layer (Ours)        & 43.2     \\
        FPI\_NN layer (Ours)        & 43.4    
    \end{tabular}
    \vspace{0.2cm}
    \caption{$F_1$ score of multi-label text classification (higher is better).}
    \label{table_mlc}
    \end{minipage}
\vspace{-0.6cm}
\end{table}
Table \ref{table_mlc} shows the $F_1$ scores. Here, DVN(Adversarial) achieves the best performance, however, it generates adversarial samples for data augmentation.
Despite their simple structures, our algorithms perform the best among those using only the training data, which confirms the effectiveness of the proposed method.
\section{Conclusion}
\label{conc}

This paper proposed a novel architecture that uses the fixed-point iteration as a layer of the neural network.
The backward FPI layer was also proposed to backpropagate the FPI layer efficiently.
We proved that both the forward and backward FPI layers are guaranteed to converge under mild conditions.
All components are highly modularized so that we can efficiently apply the FPI layer for practical tasks by only changing the structure of $g$.
Two representative cases of the FPI layer (FPI\_NN and FPI\_GD) have been introduced.
Experiments have shown that our method has advantages for several problems compared to the feedforward network. For some problems like denoising, the iterative structure of the FPI layer can be more helpful, and in some other problems, it can be used to refine the performance of an existing method. Finally, we have also shown in the multi-label classification example that the FPI layer can achieve the state-of-the-art performance with a very simple structure.

\section*{Acknowledgements}

This work was supported in part by the Next-Generation ICD Program through NRF funded by Ministry of S\&ICT [2017M3C4A7077582], and in part by the National Research Foundation of Korea(NRF) grant funded by the Korea government(MSIT) (No. 2020R1C1C1012479).

\section*{Broader Impact}

This work does not present any foreseeable societal consequence for now because it proposes theoretical ideas that can be generally applied to various deep learning structures. 
However, our method can provide a new direction for various fields of machine learning or computer vision. 
Recently, a huge portion of new techniques is developed based on deep learning in various research fields. In many cases, this leads to rewriting a large part of the traditional methodologies, because deep networks bear quite different structures from traditional algorithms. During the process, much of the conventional wisdom found in existing theories are being re-discovered based on raw datasets. This induces a high economic cost in developing new technologies. Our method can provide an alternative to this trend by incorporating the existing mechanisms in many iterative algorithms, which can reduce the development costs. There already have been many studies that combine deep learning with more complicated models such as SMPL, however, one usually has to derive the backpropagation formula separately for each method, which introduces considerable difficulties and, as a result, development costs. However, our method can be applied universally to many types of iterative algorithms, so the consilience between various models from different fields can be stimulated. Another possible consequence of the proposed method is that it might also expand the application of deep learning in non-GPU environments. As demonstrated in the experiments, the introduction of an FPI operation in a deep network can achieve similar performance with a much simpler structure, at the expense of an iterative calculation. This can be helpful for using deep networks in many resource-limited environments and may accelerate the trend of ubiquitous deep learning.

\bibliographystyle{icml}
\bibliography{refs}

\end{document}


\maketitle

\section{Figures of partial differentiation and the backward FPI layer}

The following figures show the structures of the proposed partial differentiation operator and the backward FPI layer. Here, we can see that all the operations are highly modularized, which allow multiple differentiations in a usual autograd framework without any explicit Jacobian computation.

\begin{figure}[h]
\centering
\begin{minipage}{.34\textwidth}
\centering
  \includegraphics[width=\textwidth]{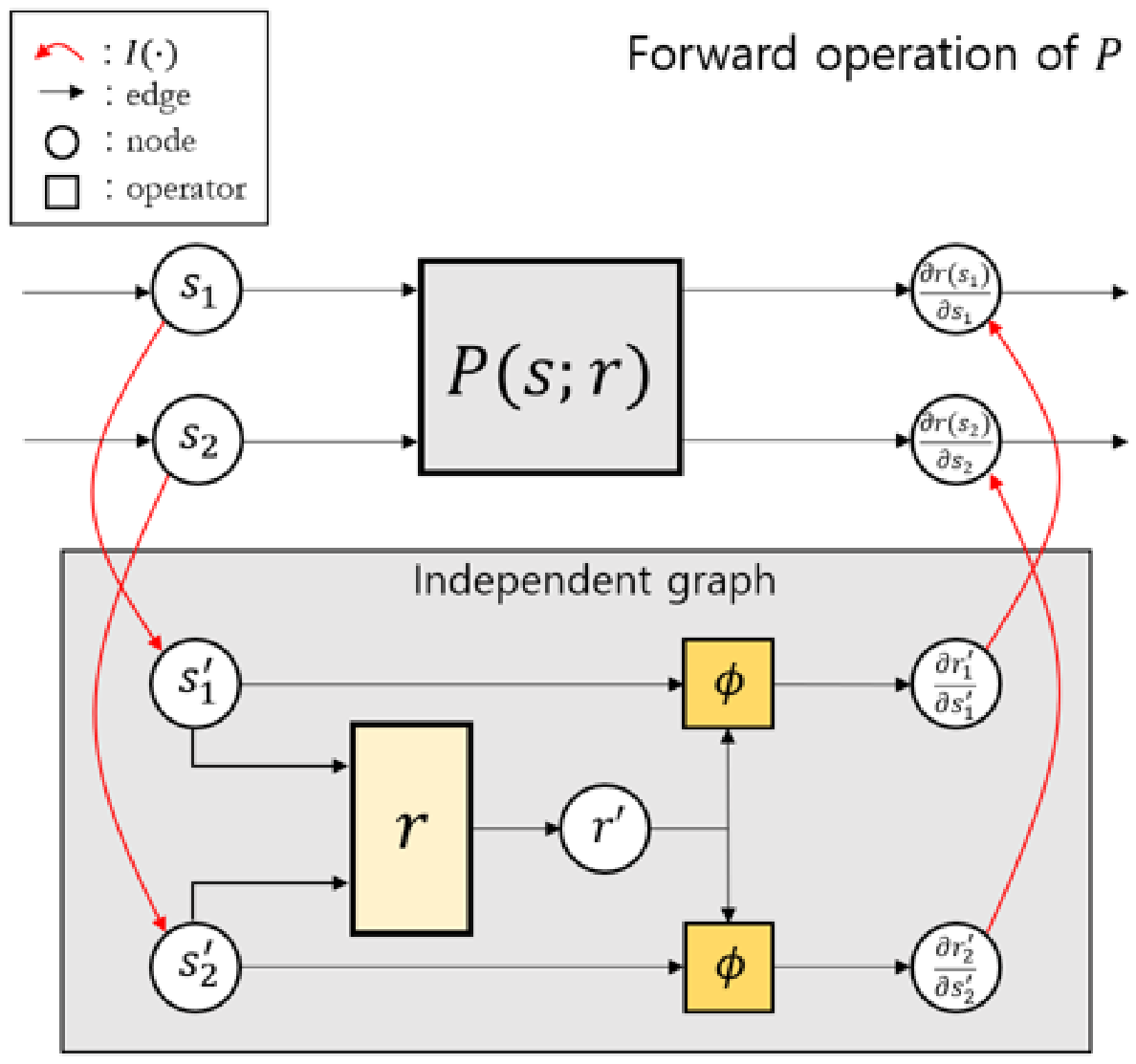}
  \caption{Forward operation of partial differentiation.}
  \label{fig_for}
\end{minipage}
\hspace{0.3cm}
\begin{minipage}{.62\textwidth}
\centering
  \includegraphics[width=\textwidth]{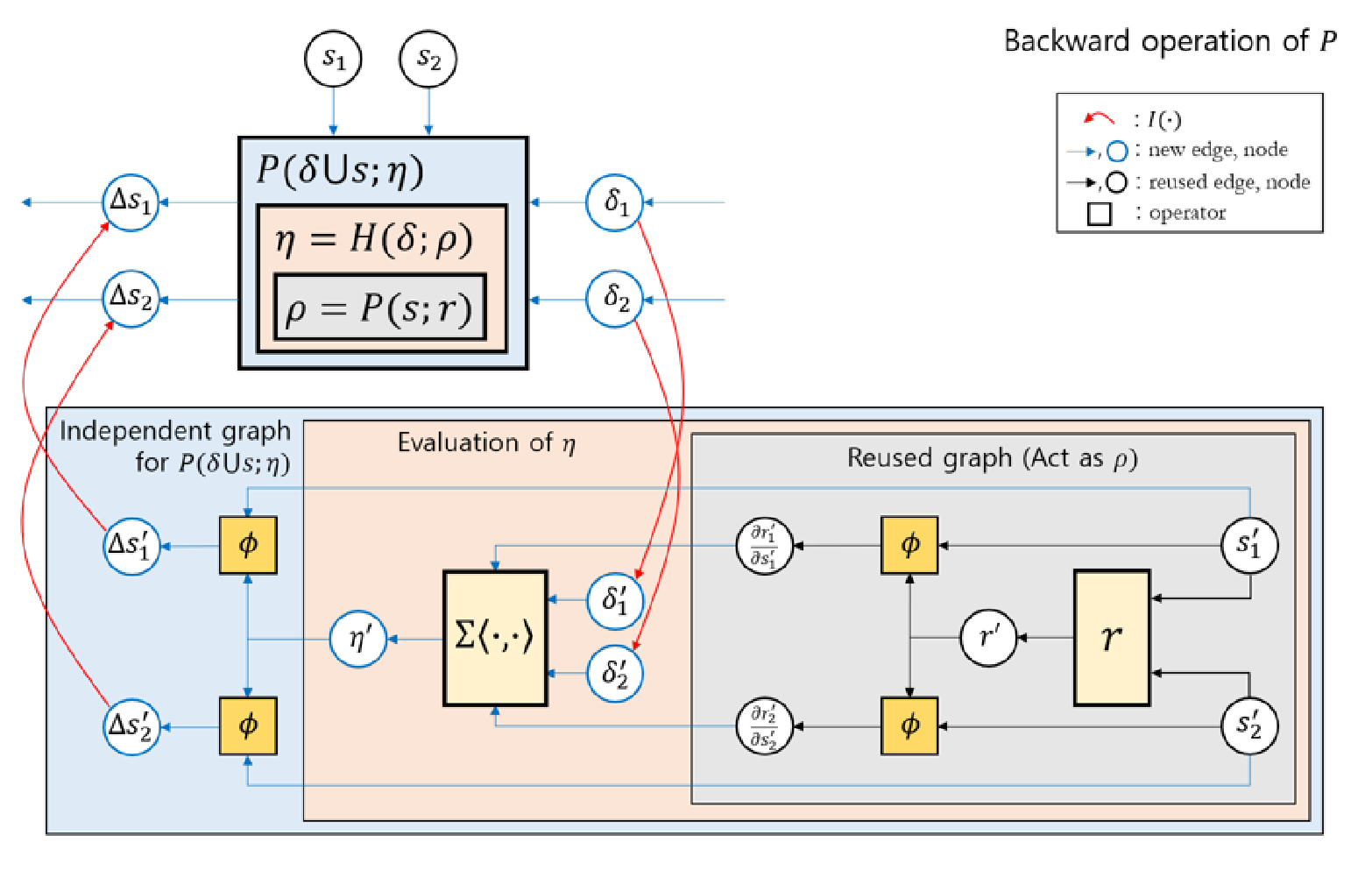}
  \caption{Backward operation of partial differentiation.}
  \label{fig_back}
\end{minipage}
\end{figure}

\begin{figure}[h]
  \centering
  \includegraphics[width=0.50\linewidth]{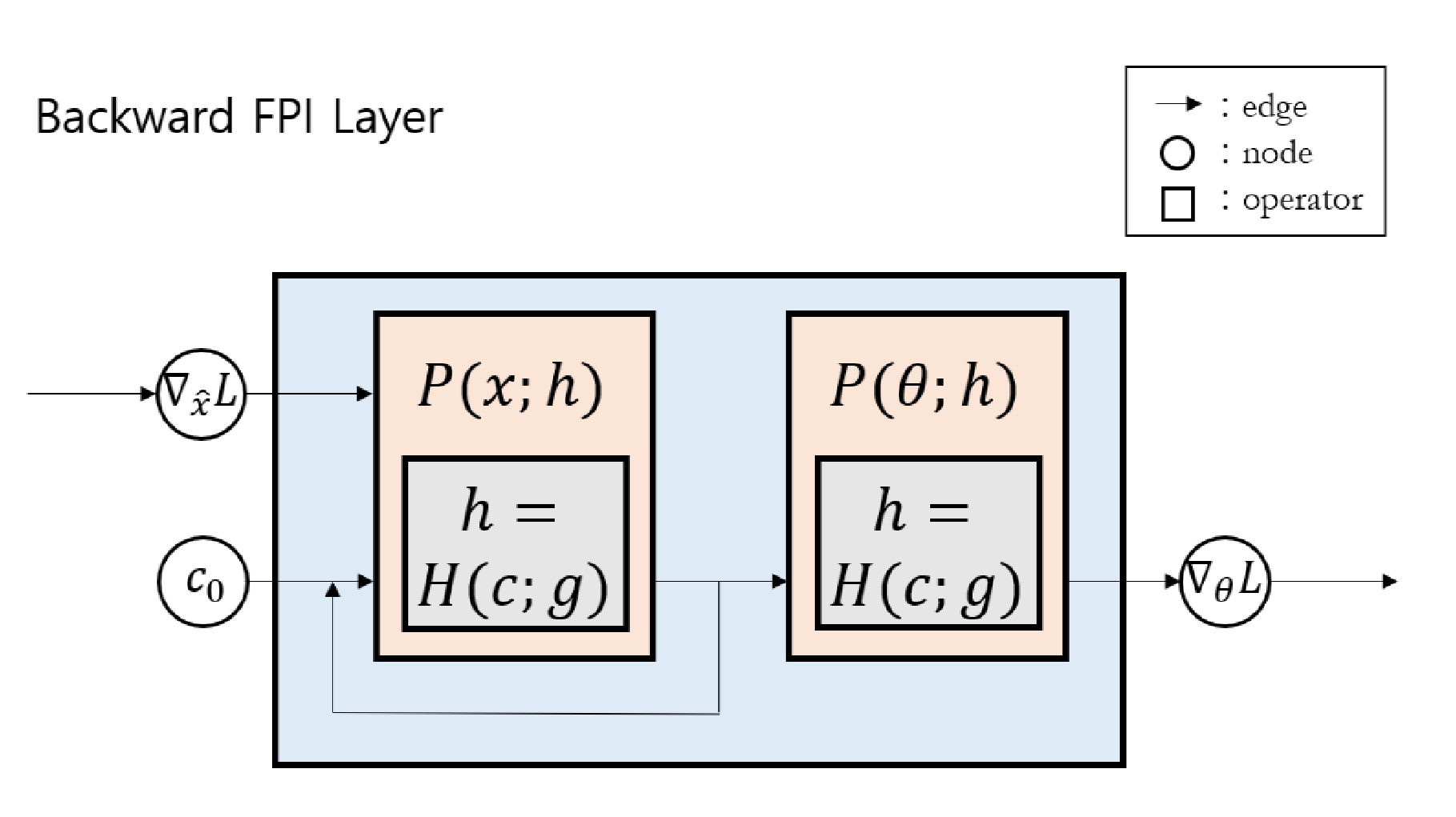}
  \caption{The backward FPI layer.}
  \label{fig_backwardFPI}
\end{figure}

\section{Implementation details}

In all experiments, we used the following criterion to determine the convergence of the FPI layer:
\begin{equation}\notag
    \beta = \cfrac{\Vert x_{n+1}-x_n\Vert^2}{\Vert x_n\Vert^2}
\end{equation}
using the $L_2$-norm. When $\beta$ went below a certain threshold, $x_{n+1}$ was considered converged and we stopped the iteration.
For all the experiments, we used two types of network modules for $g$ of the FPI layer:
\begin{enumerate}
    \item \textbf{FPI\_NN layer:} To see the full potential of the FPI layer, we tested an FPI layer with $g$ being a general (small) neural network module. In this case, $g$ can become an arbitrary function and we can explore more diverse possibilities of the FPI layer. The only issue here is that the Lipschitz constant of $g$ might not be bounded. In practice, using small initial weights for $g$ was sufficient in our empirical experience.
    \item \textbf{FPI\_GD layer:} Inspired by the energy function networks \cite{spen, icnn, dvn}, this layer performs a simple numerical optimization and yields the solution as the output of the layer. We define the energy function as a small neural network with a scalar output, and based on this energy function, $g$ is defined to be a simple gradient descent step with a fixed step size. Unlike most existing energy function networks, our version can perform backpropagation easily with the existing autograd functionalities.
\end{enumerate}

\begin{figure}[h]
  \centering
  \includegraphics[width=0.95\linewidth]{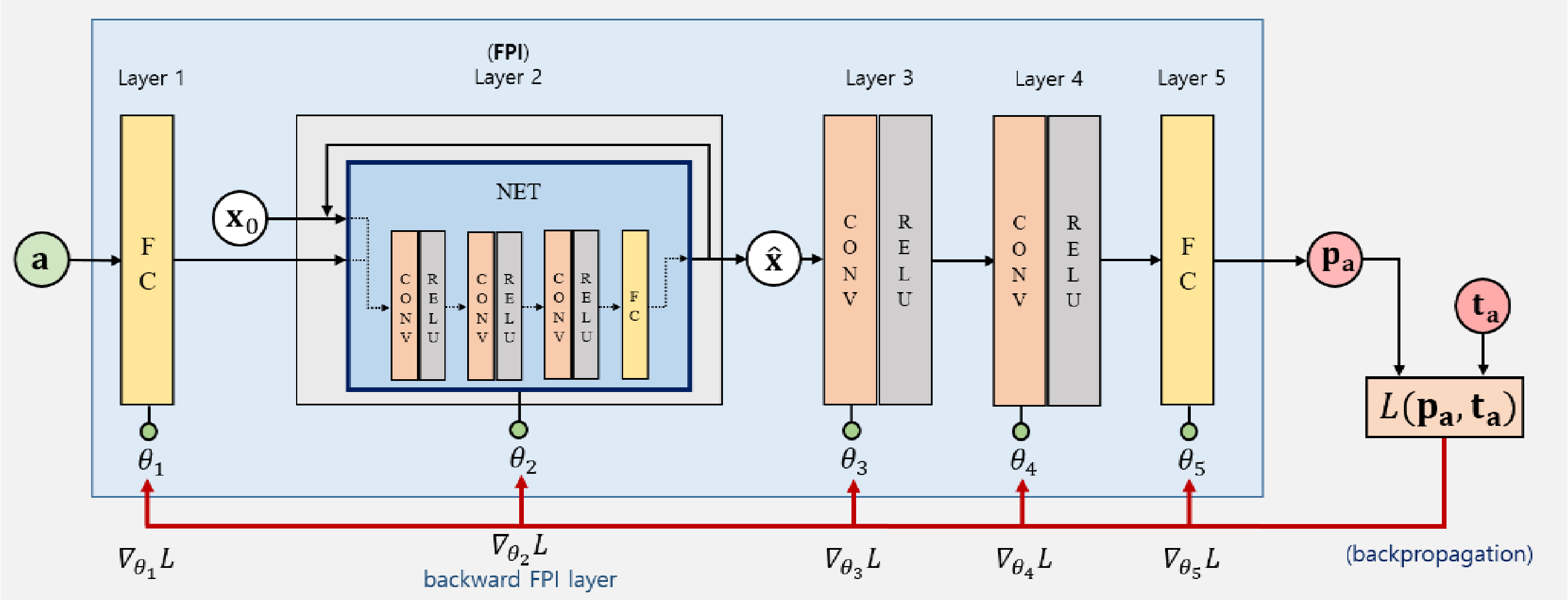}
  \caption{Example multi-layer network using FPI layer.}
  \label{fig_net}
\end{figure}

Figure \ref{fig_net} shows an example multi-layer network including an FPI\_NN layer for mini-batch size $1$. $p_a$ is the prediction for the input data $a$ and $t_a$ is the ground truth.
As can be seen in the figure, the FPI layer composed of a neural network module.
When passing the FPI layer during the backpropagation process, backward FPI layer is used to compute $\nabla_{\theta_2}L$.

For the experiments with image inputs, such as image denoising and optical flow, only the FPI\_NN layer was tested.
For the energy function network of the FPI\_GD layer to have a scalar output, we attached a mean squared layer at the end of the network.
For the multi-label classification, the performance was evaluated for both the FPI\_NN and FPI\_GD.
Note that, for all the above layers, the fixed-point iteration variable $x$ was concatenated with the layer input $z$, and is passed onto $g$ (e.g., for vector inputs, $g(x, z;\theta) = g([x^T \,\, z^T]^T; \theta)$). Accordingly, the size of the input of $g$ was bigger than that of the output. $x_0$ was either a zero vector or a zero matrix in all the experiments.

All the training was performed using the Adam optimizer with learning rate $10^{-3}$, and no weight decay was used.
In the following experiments, we used the ReLU activation function most of the time. Although this does not exactly align with the assumption in the paper, we used it anyway as in many practices of deep learning and confirmed that the FPI layer still performs nicely. 

\subsection{Image denoising}
The first 500 images of the flying chair dataset \cite{flownet} were cropped to $180\times180$ around the center, of which the former 400 images were used to train the networks and the latter 100 were used as test samples.
All images were converted to gray scale.

\begin{figure}[h]
\centering
    \includegraphics[width=0.5\textwidth]{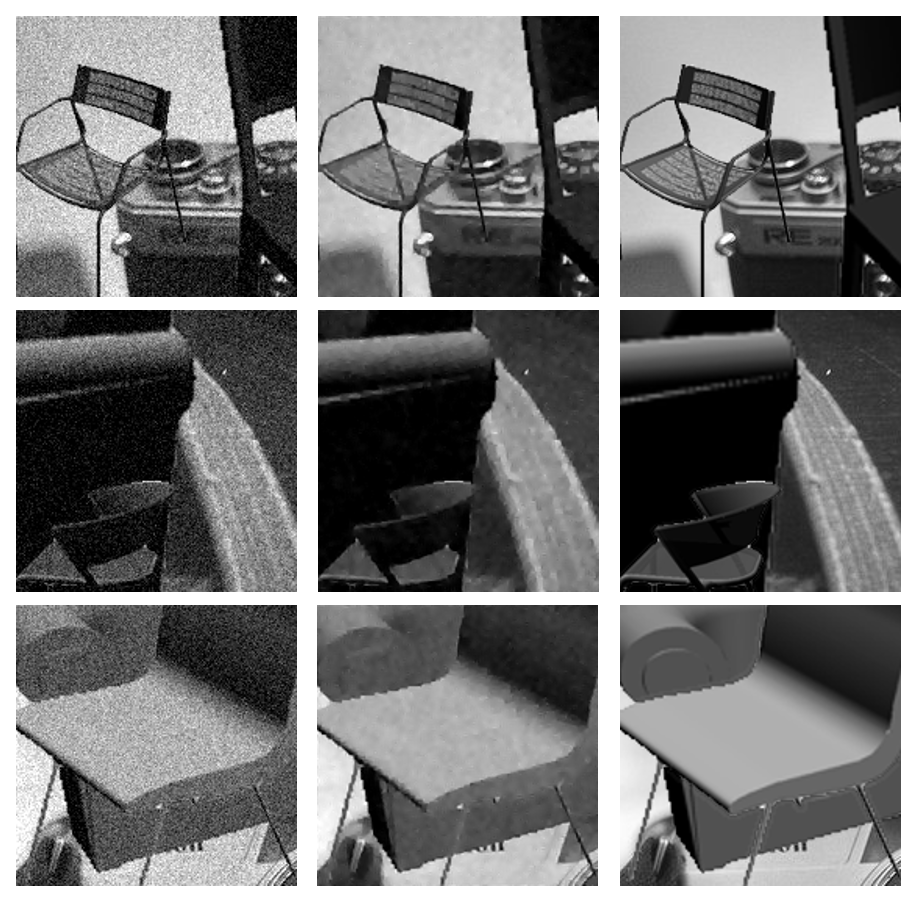}
    \caption{Image denoising examples. The left column shows the noisy input and the right column is the ground truth. The denoised images are shown in the middle.}
    \label{fig_dn}
\end{figure}

All the network modules consisted of two 2D convolution layers with 32 intermediate channels and a ReLU activation after the first convolution layer in the proposed method, and the baseline (non-FPI) feedforward network also shared the same structure.
The channel sizes of the network's input and output were both one since the input and the output were gray-scale images.
All the models were trained for 20 epochs, and the final results were reported based on the best epoch for each method.
Gradient clamping with max norm 0.1 was used to prevent the divergence of the FPI layers and The convergence threshold of the FPI layer was set to $10^{-7}$.
The initial values of the network weights were set ten times smaller than the default initialization of PyTorch \cite{xavier}.
Figure \ref{fig_dn} shows the image denoising examples.

\subsection{Optical flow}
We used the FlowNetS \cite{flownet} structure for this experiment.
As in \cite{flownet}, we tested the performance on the flying chair dataset with the same training/test split.
The number of channels in FlownetS starts at 6 and increases to 1024 using 10 conv layers, which are followed by several deconv layers.
We attached an FPI layer at the end of the FlowNetS, where $g$ consists of one conv layer with four input and output channels and one deconv layer with four input and two output channels.
The strides of the conv and deconv layers were both set to two for downsampling and upsampling.
The final output of the proposed model was the summation of the output of the FPI layer and that of the original FlowNetS. 
Note that both the FlowNetS and the FPI layer were trained end-to-end in this experiment.
The convergence threshold of the FPI layer was set to $10^{-13}$.
The initial values of the network weights was set 100 times smaller than the default initialization of PyTorch.
All other training settings were identical to the original FlowNet.
Note that, for this experiment, the performance was worse than that of the original FlowNetS when a (non-FPI) feedforward module of the same structure was attached at the end of FlowNetS, and thus this result is omitted from the experimental results.

\subsection{Multi-label classification}
For this experiment, all the training settings of the proposed methods were the same as the other compared algorithms, except for the network structure.
The network structure of FPI\_NN was composed of two fully-connected (FC) layers with 512 hidden nodes, a ReLU activation after the first FC layer, and
an additional sigmoid layer after the second FC layer to normalize the output between zero and one. 
In this case, the convergence threshold was $10^{-8}$.
For FPI\_GD, the energy function had only one FC layer with ReLU activation and a mean squared term to have a scalar output.
Here, the number of hidden nodes were also 512, and an additional sigmoid layer was added after the FPI\_GD layer.
We fixed the step size to $1.0$ for the gradient descent in FPI\_GD and used a different convergence criterion as follows:
\begin{equation}\notag
    \beta' = \Vert x_{n+1}-x_n\Vert^2
\end{equation}
where the convergence threshold was $10^{-4}$.
The sizes of both the networks' inputs and outputs were 1836 and 159, respectively, which is the same as the numbers of indicators and labels.

\section{Additional experiments: a constrained problem}

Here, we show the feasibility of our algorithm by learning a constrained optimization problem ($\min_{ x}\left\Vert x-a\right\Vert^2$) with a box constraint ($-\mathbf{1}\le x\le\mathbf{1}$).
The goal of the problem is to learn the functional relation based on training samples $(a, t)$, where $t$ is the ground truth solution of the problem. We used single FPI layer networks for this problem.

\begin{figure}[h]
  \centering
  \includegraphics[width=0.45\linewidth]{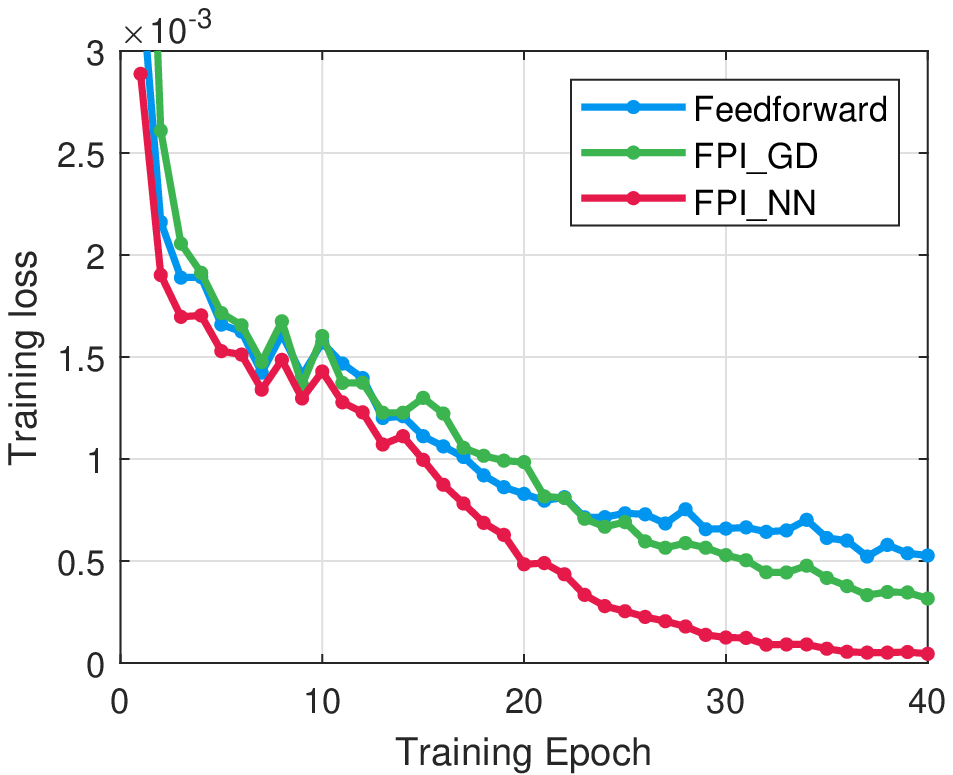}
  \includegraphics[width=0.45\linewidth]{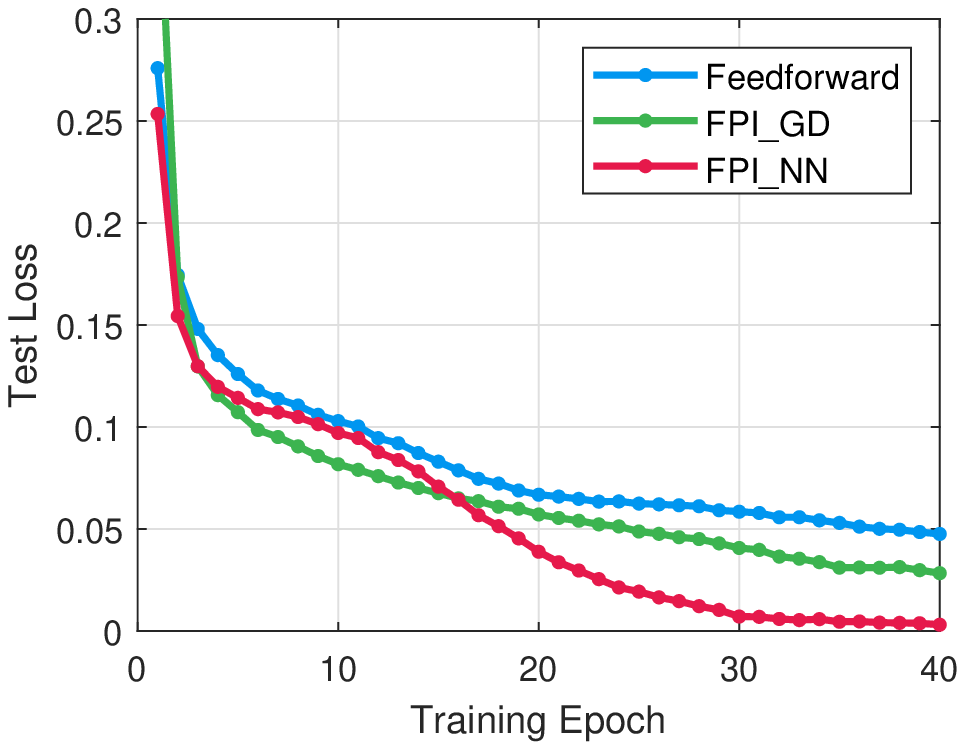}
  \caption{Training and test loss per epoch in constrained problem.}
  \label{fig_toy}
\end{figure}

The performance was evaluated for the FPI\_NN network, the FPI\_GD network, and a non-FPI network which has the same structure with $g$ of FPI\_NN. The structures of $g$ of FPI\_NN and the energey function of FPI\_GD were both linear-ReLU-linear. The dimension of $a$, $x$, and the number of hidden nodes were 10, 10, and 32, respectively. We randomly generated $a$ from zero-mean Gaussian distributions.
10,000 training samples were generated with $\sigma = 2$ and 1,000 test samples with $\sigma = 1$.
The convergence threshold was set to $10^{-6}$ for both FPI\_NN and FPI\_GD layers and the step size of the gradient descent in FPI\_GD was fixed to $0.01$.
All the models were trained for 40 epochs with batch size 100, and the training and test losses (MSE) were reported.

Figure \ref{fig_toy} shows the train and test losses per epoch. Here, we can see that FPI\_NN outperforms the other networks for both the train and the test losses.

\section{Additional lemmas for the convergence of backward FPI layer}

Here, we use an arbitrary norm metric for all the vector and matrix norms.
The following lemma holds for vectors $x$ and $b$, matrix $A$, and scalar $k < 1$.





\textbf{Lemma 1.}
\textit{If the matrix norm $\Vert A\Vert < 1$, then the linear transform by weight $A$, i.e., $f(x) = Ax+b$, is a contraction mapping.}

\begin{proof}

\begin{equation}\notag
\begin{split}
    & \cfrac{\Vert f(x_1)-f(x_2)\Vert}{\Vert x_1-x_2\Vert}
    = \cfrac{\Vert Ax_1-Ax_2\Vert}{\Vert x_1-x_2\Vert} = \cfrac{\Vert A(x_1-x_2)\Vert}{\Vert x_1-x_2\Vert} \\
    &\leq \cfrac{\Vert A\Vert\cdot\Vert x_1-x_2\Vert}{\Vert x_1-x_2\Vert} = \Vert A\Vert \leq k < 1
\end{split}
\end{equation}

By the definition, $f$ is a contraction mapping.

\end{proof}

In section 3.4, 
$\Vert J_g(\hat{x})\Vert \leq k < 1$, which means that the linear transform with weight matrix $J_g(\hat{x})$ is a contraction mapping by \textit{Lemma 1}. Therefore, (10) of the backward FPI in the main paper is a contraction mapping.





\bibliographystyle{icml2020}
\bibliography{References}